%% file: main.tex
\documentclass[10pt,twocolumn,letterpaper]{article}

\input{latex/macros.tex}
\usepackage[breaklinks=true,bookmarks=false]{hyperref}

\cvprfinalcopy 

\def\cvprPaperID{2843} 

\ifcvprfinal\fi
\begin{document}

\title{Blind Visual Motif Removal from a Single Image}
\author{Amir Hertz\qquad Sharon Fogel\qquad Rana Hanocka\qquad Raja Giryes\qquad Daniel Cohen-Or \vspace*{0.2cm} \\ Tel Aviv University}
\twocolumn[{
\renewcommand\twocolumn[1][]{#1}
\maketitle
\begin{center}
    \centering
    \vspace*{-0.5cm}
    \includegraphics[width=\textwidth]{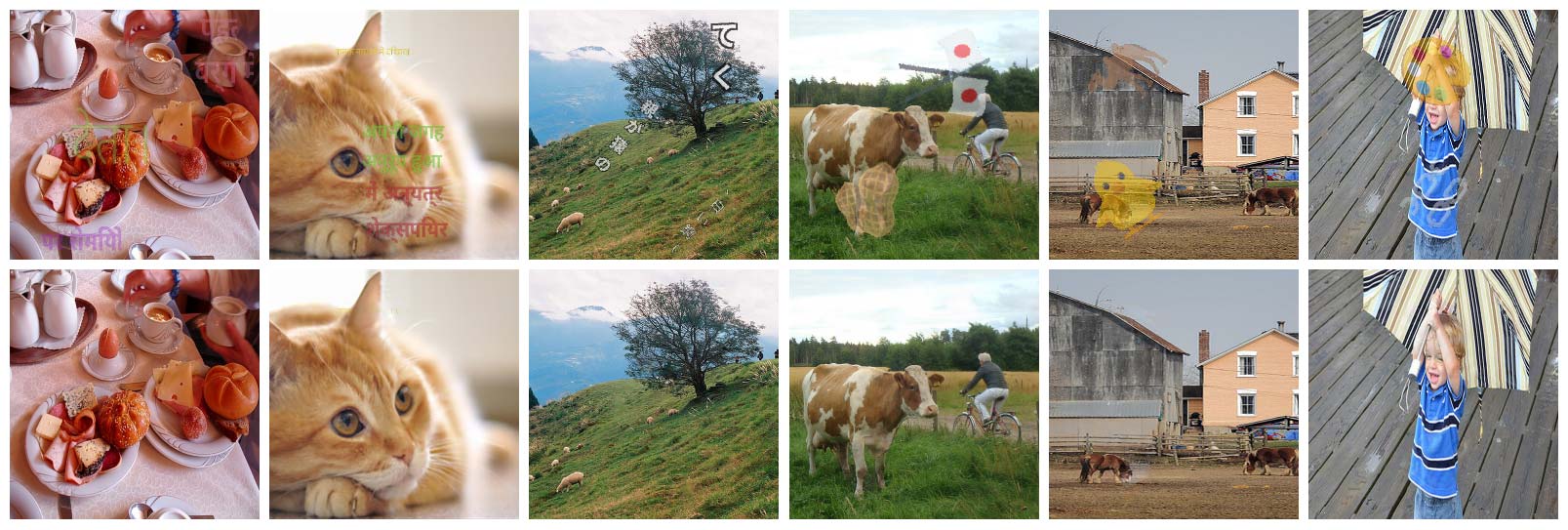}
    \vspace*{-0.5cm}
    \captionof{figure}{Blind visual motif removal results on images unseen during training. Top: test images embedded with semi-transparent motifs. Bottom: our reconstructed results. Our network was trained on Latin characters, yet successfully identifies and removes the Hindi and Japanese characters (left three images). Similarly, the overlaid visual motifs on the right three images differ semantically from the motifs used during training.}
\label{fig:teaser}
\end{center}
}]
\thispagestyle{empty}


\input{latex/abstract.tex}

 \input{latex/introduction.tex}

\input{latex/related.tex}

 \input{latex/method.tex}

 \input{latex/results.tex}
 \input{latex/conclusion.tex}
{\small
\bibliographystyle{ieee}
\bibliography{egbib}
}

\end{document}

%% file: latex/macros.tex
\usepackage{cvpr}
\usepackage{times}
\usepackage{epsfig}
\usepackage{graphicx}
\usepackage{amsmath}
\usepackage{amssymb}

\long\def\ignorethis#1{}

\usepackage{url}

\usepackage{xfrac}
\usepackage{newtxmath}
\usepackage{calrsfs}
\DeclareMathAlphabet{\pazocal}{OMS}{zplm}{m}{n}

\usepackage{booktabs}

\usepackage{soul}
\soulregister\ref{7}
\soulregister\cite{7}
\soulregister\refFig{7}

\usepackage{color}
\definecolor{gray}{rgb}{0.35,0.35,0.35}
\definecolor{blue}{rgb}{0,0,1}
\definecolor{white}{rgb}{1,1,1}
\definecolor{peach}{rgb}{0.97,0.59,0.35}
\definecolor{aquamarine}{rgb}{0.2,0.504,0.467}

\usepackage{booktabs}

\usepackage{epsfig}
\usepackage{epstopdf}
\usepackage[position=top]{subfig}
\usepackage{multirow}
\usepackage{array}

\usepackage{xcolor}

\newbox\jsavebox

\newcommand{\transition}{segment}
\newcommand{\transitions}{segments}
\newcommand{\Bm}{\alpha}
\newcommand{\emm}{\circ}

%% file: latex/abstract.tex
\begin{abstract}
Many images shared over the web include overlaid objects, or \emph{visual motifs}, such as text, symbols or drawings, which add a description or decoration to the image.
For example, decorative text that specifies where the image was taken, repeatedly appears across a variety of different images. 
Often, the reoccurring visual motif, is semantically similar, yet, differs in location, style and content (\emph{e.g.,} text placement, font and letters).
This work proposes a deep learning based technique for \emph{blind} removal of such objects. In the blind setting, the location and exact geometry of the motif are unknown. Our approach simultaneously estimates which pixels contain the visual motif, and synthesizes the underlying latent image. It is applied to a single input image, without any user assistance in specifying the location of the motif, achieving state-of-the-art results for blind removal of both opaque and semi-transparent visual motifs. 

\end{abstract}

%% file: latex/introduction.tex
\section{Introduction}

Images shared over the web often contain overlaid objects, or \emph{visual motifs}, such as text, symbols or drawings, which can serve a variety of purposes.
%
For example, adding university emblems or team logos can instill an image with pride, or, text with temperature information can add extra information to an image.
In a subset of cases, motifs are used as a protection of image ownership, \emph{e.g.,} with digital watermarks. 
However, in many cases, one may want to remove these visual motifs and reconstruct the original image without the existing occlusions. 


The removal of these visual motifs and the recovery of a pristine image can be an extremely challenging task. The structure, size and location of these objects varies between different images, making them difficult to detect without user guidance or assumptions about the underlying image.
Previous methods have relied on information about the location of the corrupted pixels to be restored \cite{pei2006novel, park2012identigram, ren2015shepard, zoran2011learning}.
Dekel et al. \cite{dekel2017effectiveness} remove watermarks using large image collections, which contain the same watermark, as well as some minimal user guidance about the watermark location. 

We present a method for completely \emph{blind} visual motif removal. In the blind setting, the exact location, structure and size of these motifs is unknown.
%
%
The generalization ability of our network is demonstrated by removing visual motifs that are not seen during training, \emph{e.g.,} watermark removal from real-world images. See Figures 1 and 8 for examples of visual motif removal.


Unlike previous approaches, our strategy does not require multiple images with the same object to be removed, or the exact location of the motif pixels. 
Our technique exploits the ability of deep neural networks to generalize from samples seen during training to unseen samples during test time, thereby enabling the removal of novel (\emph{i.e.,} unseen) motifs.
The network learns to separate the visual motif from the image, by first estimating which pixels contain the motif and subsequently reconstructing the latent image. 
In addition to estimating the binary motif matte and reconstructing the latent image, we also reconstruct the visual motif. 

Decomposing the input image into both the visual motif and the background image enables the network to create a better reconstruction of the image.
In our experiments, we demonstrate that images corrupted by different types of visual motifs such as text or semi-transparent drawings can be recovered successfully.

%% file: latex/related.tex
\section{Related work}

While blind removal of visual motifs is not addressed directly in the literature, there are related works in the watermark removal and blind image inpainting fields. 

{\bf Watermark removal.} 
Digital watermarks are commonly used to identify the copyright of images to prevent users from using images available online without consent.
Though many techniques for creating watermarks are available \cite{tao2014robust}, the visible watermarking process is usually composed of embedding a logo or text with varied opacity into a target image.

Watermark attack methods, \emph{e.g.,}~\cite{voloshynovskiy2001attacks}, strive to accurately remove watermarks from a host image. The robustness of a watermark to an attack is essential for protecting image copyrights. 
Watermark removal methods have been researched to test and improve the resilience of watermarks.
Removing a watermark typically involves two steps: watermark detection and reconstruction of the underlying image content. 

Most watermark removal methods try to reconstruct a single watermarked image relying on user guidance to identify the location of the watermark. 
Huang and Wu \cite{huang2004attacking} obtain the watermark location from the user and then use image inpainting to restore the latent image. 
Pei and Zeng \cite{pei2006novel} use independent component analysis (ICA) to separate the source image from the watermark. Park et al. \cite{park2012identigram} find a color space transformation that separates the source image from the watermark. More recently, 
Dekel et al. \cite{dekel2017effectiveness} assume the existence of a collection of images with the same watermark to estimate and remove it. 
Contrary to these methods, our method does not require any user intervention and the watermark can be removed using a single image only.


{\bf Image inpainting.} Image inpainting is the process of restoring or completing portions of an image \cite{ballester2001filling, barnes2009patchmatch, bertalmio2000image, criminisi2004region, perez2003poisson}. Most image inpainting approaches are \emph{non-blind}, since they assume that the binary mask indicating the pixels to be restored is given \cite{gilbert2018disentangling,iizuka2017globally, Tirer19Image,Yu10Solving,yu2018generative}.

Some works consider the case of \emph{blind} inpainting, where the mask of the pixels to be restored should be detected as well. Zoran et al. \cite{zoran2011learning} present a GMM based model for estimating the binary mask and restoring the latent image.
Other approaches use deep neural networks with both blind and non-blind settings that can be used recover the latent image \cite{liu2017deep,ren2015shepard}.
Clearly, in all cases the performance degrades when the inpainting is performed blindly, without the mask of the missing pixels.



It is possible to use blind inpainting for watermark removal, but this would not take into account the latent image information provided by the watermark semi-transparency. In this sense, it would be inferior to methods which use this available information (\emph{e.g.,} \cite{dekel2017effectiveness}).


{\bf Image restoration.}
Other previous works related to the removal of visual motifs include the removal of raindrops \cite{eigen2013restoring,Kligvasser18xUnit,qian2018attentive,zhang2017image}, distracting objects \cite{fried2015finding}, and reflections \cite{fan2017generic, Yang2018Seeing, Zhang2018SingleIR} from images.
Our work takes inspiration from a reflection separation work \cite{lee2018generative}, which separates an image with a reflection into two clear images. 
Contrary to this method, we use the separation to remove visual motifs from an image. The elements removed by our method can be of different sizes. They can be very small compared to the entire image and therefore hard to detect.

%% file: latex/method.tex
\section{Method}
\input{figures/arch/arch_diagram.tex}
\input{figures/arch/unet_diagram.tex}

Removing a visual motif from a single image is a difficult task to achieve without prior knowledge about the motif or the image. Our proposed approach tackles this problem using a convolutional neural network (CNN) trained to remove visual motifs embedded in an image. We train the network using a synthesized dataset of images with visual motifs (see Figure \ref{fig:batch_example}). 

Our network learns to separate the visual motif from the image, by estimating the visual motif matte and reconstructing the latent image. During training, the loss computation uses the input image and the visual motif as ground-truth to train an encoder and decoder networks. Our network encodes the corrupted image into a latent representation, which is decoded by three parallel decoder branches: one for estimating the latent image, motif matte and motif image. The final image is generated by using the estimated motif matte to select pixels from either the input image or the reconstructed image. See Figure~\ref{fig:arch_diagram} for an overview.

\subsection{Motif embedding}
Matting a motif ($Vm$) onto an image ($Im$) can be obtained by:
\begin{equation}
{Cr} = {\Bm{}} \emm {Vm} + \left(1 - {\Bm{}} \right) \emm {Im},
\end{equation}
where $Cr$ is the synthesized corrupted image, and $\Bm{}$ is the spatially varying transparency, or \emph{alpha matte}.
The alpha matte contains a scalar $\alpha_i$ at pixel location $i$, such that $\alpha_i\in\left(0, 1\right)$ inside the motif region and $0$ elsewhere.
In the case where the corrupted image contains a completely opaque motif (\emph{i.e.,} when $\alpha_i\ = 1$ for all motif pixels), recovering the latent image is equivalent to image inpainting.
Following the work of \cite{dekel2017effectiveness}, we also experiment with more challenging motifs by applying a spatial perturbation to motif pixels before matting.
For recovering the latent image $im$, both $\Bm{}$ and the visual motif $Vm$ must be known. Then it is simply given by
\begin{equation}
{Im} =  \frac{{Cr} - {\Bm{}} \emm {Vm}}{1 - {\Bm{}}}.
\label{eq:im_rec}
\end{equation}
Similarly, the recovery of the visual motif requires the latent image $Im$ in addition to the $\Bm{}$ matte information.

\subsection{Blind visual motif removal}

Our baseline network is composed of one encoder and two decoder branches: one for estimating the original background image $\widehat{Im}$, and another that reconstructs the visual motif mask $\widehat{Ma}$. 

The final image is reconstructed by replacing all the pixels in the estimated visual motif region in the original corrupted image $Cr$ with the corresponding pixels from the reconstructed image $\widehat{Im}$. The final resulting image is given by
\begin{equation}
Im_{final} = (1-\widehat{Ma}) \emm Cr + \widehat{Ma} \emm \widehat{Im},
\end{equation}
where $\emm$ denotes element-wise multiplication. 

The loss term of the baseline network is composed of two reconstruction losses: one for the estimated visual motif mask (denoted $L_{mask}$), and another for the background image in the corrupted area (denoted $ L_{im}$). The total loss for the baseline network is
\begin{equation}
L = L_{mask} + \lambda L_{im}.
\end{equation}
For the mask loss $L_{mask}$, we use binary cross entropy between the estimated $\widehat{Ma}$ and ground-truth $Ma$ motif masks
\begin{align}
& L_{mask} = \\ \nonumber
& \hspace{-0.05in} -\frac{1}{W\cdot H}\mathop{\sum\sum}_{\substack{0<i<H \\ 0<j<W}} Ma_{ij} \log{\widehat{Ma_{ij}}} + (1-Ma_{ij} ) \log{\left(1-\widehat{Ma_{ij}}\right),}
\end{align}
where $W$ is the width and $H$ is the height of the image.

For the image restoration loss $ L_{im}$, we use the $\ell_1$ distance between the recovered image $\widehat{Im}$ and the original image $Im$ in the foreground mask pixels:
\begin{equation}
L_{im} = \frac{1}{|Ma|} \cdot \ell_1\left(Pix, \widehat{Pix} \right),
\end{equation}
where $|Ma|$ is the size of the mask (in pixels); 
$Pix$ and $\widehat{Pix}$ represent the values in the region of the visual motif, of the original background image $Im$ and the estimated image $\widehat{Im}$ respectively. These are given by:
\begin{equation}
Pix = Im \emm Ma  \hspace{0.5cm} \widehat{Pix} = \widehat{Im} \emm Ma.
\end{equation}
Notice that we use the original mask ${Ma}$ (and not  ${\widehat{Ma}}$) in $\widehat{Pix}$ for calculating the $L_{im}$ loss during training. We do this in order to separate between the loss terms of the two decoders (that generate $\widehat{Im}$ and  ${\widehat{Ma}}$).
Otherwise, $L_{im}$ affects the mask reconstruction decoder such that it tends to converge to local minima (an empty motif mask).

Naturally, in some cases, the goal would be to reconstruct the visual motif. For such cases, we add another decoder branch for inferring the original visual motif. This network outputs the image $\widehat{Vm}$ and has shared weights with the image decoder in the first few layers (see Figure~\ref{fig:arch_diagram}). Later, in the ablation study in Section \ref{sec:abl}, we show that the insertion of the additional branch does not only enable the recovery of the visual motif, but also leads to a small improvement in the quality of the reconstruction of the background image.

The addition of the visual motif branch adds the loss term
\begin{equation}
L_{Vm} = \frac{1}{|Ma|} \ell_1\left(Vm, \widehat{\widehat{Vm}} \right),
\end{equation}
where $Vm$ and $\widehat{\widehat{Vm}}$ represent the ground truth visual motif before blending and the generated visual motif given by:
\begin{equation}
 \widehat{\widehat{Vm}} = \widehat{Vm} \emm Ma,
\end{equation}
where $\widehat{Vm}$ is the output of the bottom branch in Figure~\ref{fig:arch_diagram}.
All the above lead to the following final loss term:
\begin{equation}
L = L_{mask} + \lambda \left( L_{im} + L_{Vm}\right).
\end{equation}

The shared weights at the lower levels of the image and visual motif decoders emerges from the simple observation in equation \ref{eq:im_rec}, where recovering one helps in recovering the other.


\subsection{Network architecture}

Each sub-network is a fully convolutional deep neural network, based on the U-Net architecture \cite{ronneberger2015u}: the feature maps from decoder layer are combined with the corresponding encoder layer features maps at the same resolution.

Figure \ref{fig:unet_diagram} shows the architecture details.
The encoder and decoders contain five main \transitions{}. In the encoder, each \transition{} increases the number of channels by a factor of two. It includes $3 \times 3$ convolutional layers, batch normalization~\cite{ioffe2015batch} and the ReLU  activation unit~\cite{nair2010rectified}. The spatial size of the input is reduced using max pooling with stride-$2$. 

The decoders perform an \emph{inverse operation}. Each decode \transition{} expands the spatial size of the input and reduces the number of channels by a factor of two. It starts with a $3 \times 3$ transposed convolutional layer with stride-$2$. Afterwards, an additional convolutional layer is applied on both the previous output and the corresponding encoded data. Batch normalization and ReLU are also used.
After the last \transition{} of the decoder, the final output is generated by applying a $1 \times 1$ convolutional layer followed by a sigmoid activation for the mask decoder and a $\tanh$ activation for the image and visual motif decoders. 

Notice that in both the encoder and decoders, three residual blocks are used in each \transition{}. This expands the receptive field and improves the quality of the recovered image (see Section~\ref{sec:abl}).

%% file: figures/arch/arch_diagram.tex
\begin{figure*}
\includegraphics[width=\textwidth]{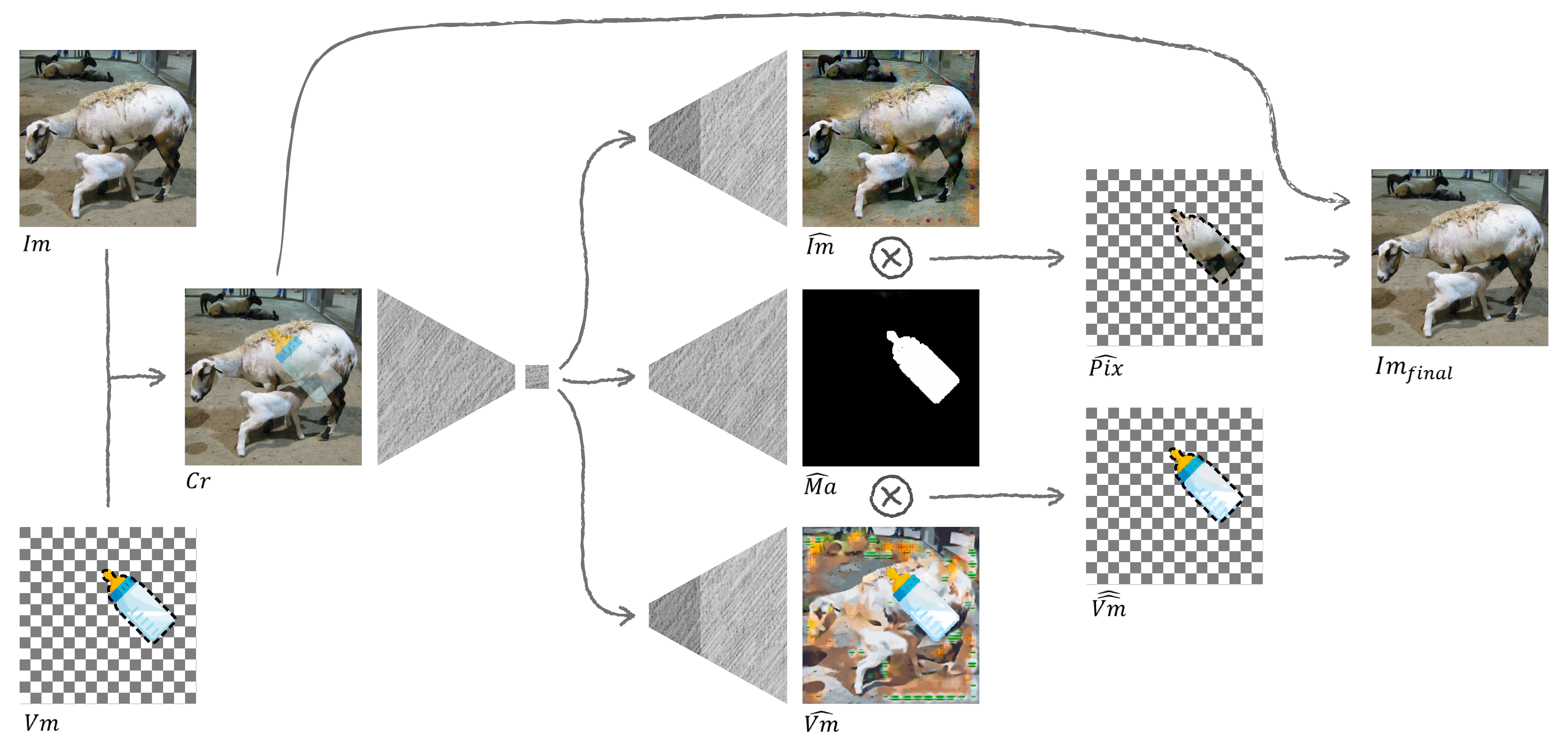}
\caption{{\em Method overview.}
The network consists of one encoder and three decoders. The top and bottom decoder branches reconstruct the background image and the overlaid visual motif, respectively. These decoders share weights in the first few layers (marked by darker gray color). The middle branch estimates the mask of the visual motif. 
The final output is generated by using the mask to select pixels from either the input image or the reconstructed image.
}
\label{fig:arch_diagram}
\end{figure*}

%% file: figures/arch/unet_diagram.tex
\begin{figure}
\subfloat[Encoder-decoder main levels.]{\includegraphics[width=\columnwidth]{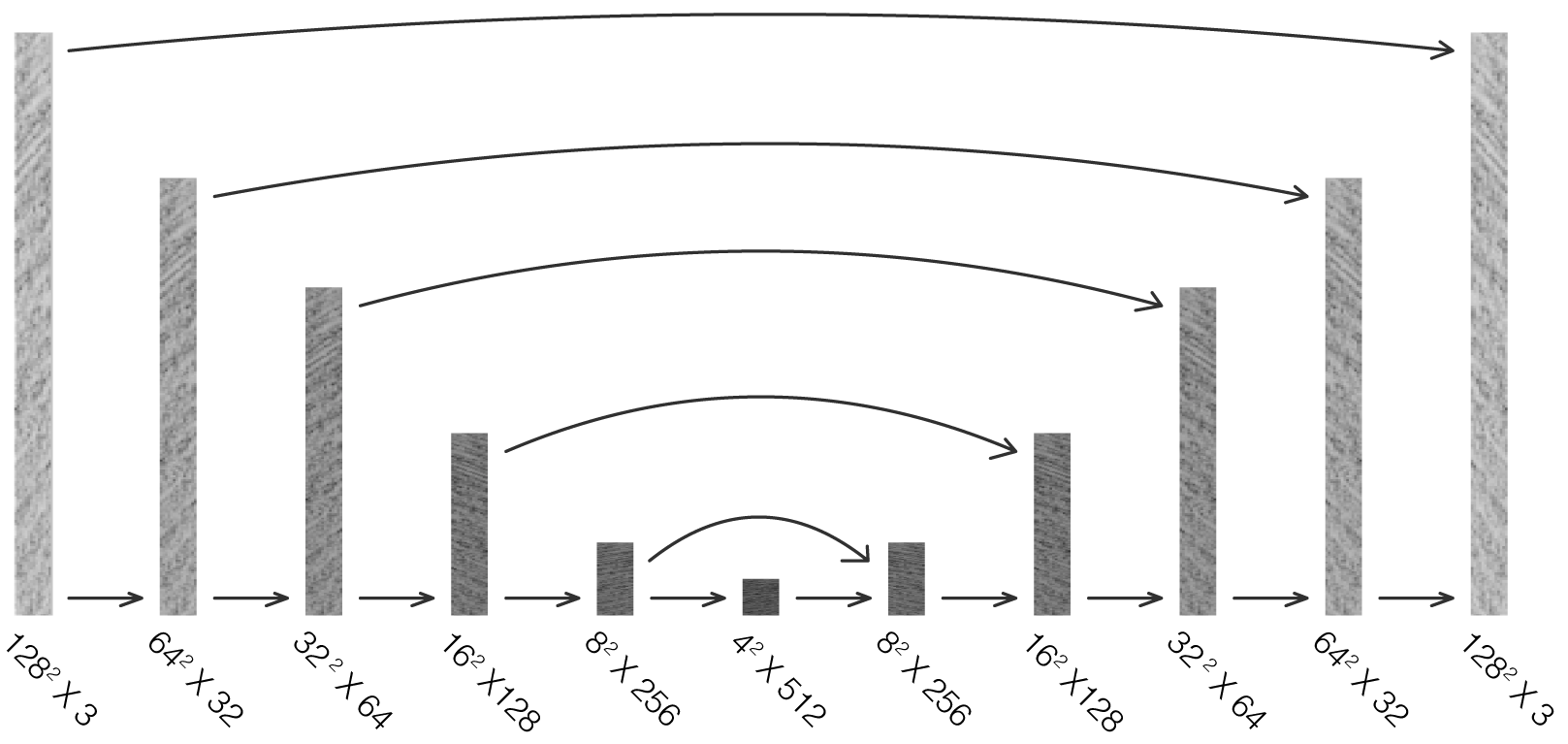}}\\
\subfloat[Detailed view of an encode \transition{}.\vspace*{-0.2cm}]{\includegraphics[width=\columnwidth]{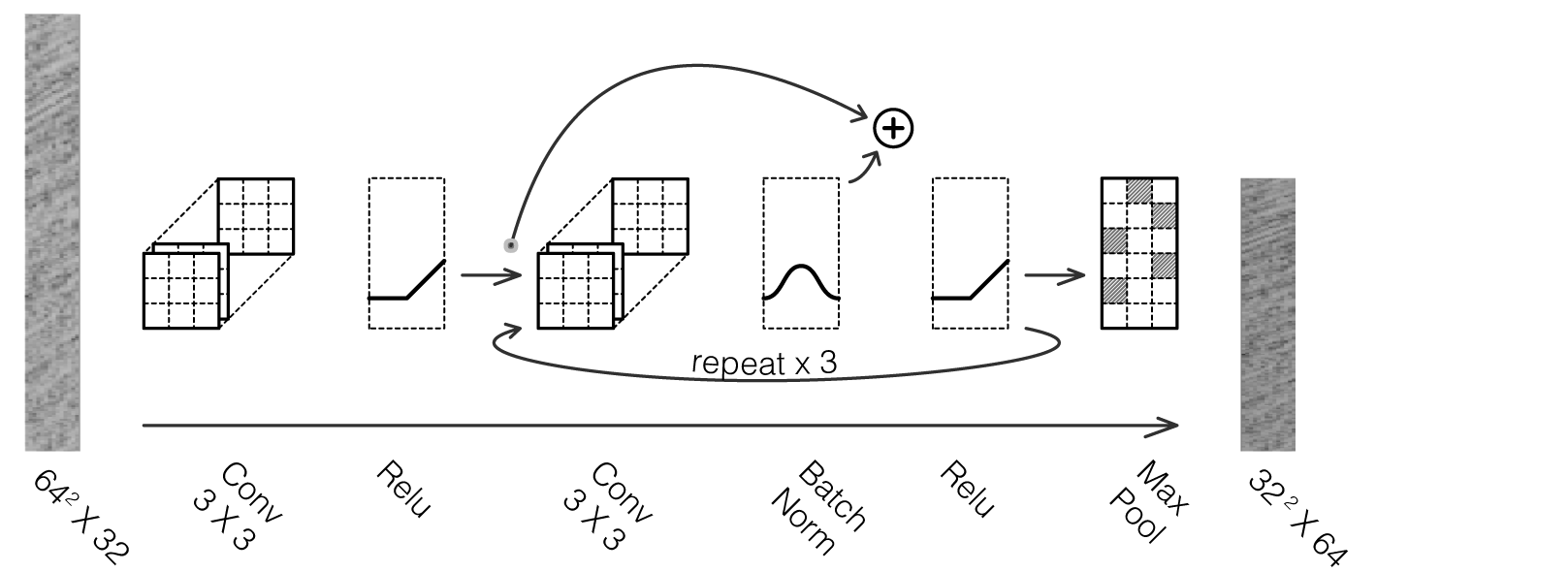}}\\
\subfloat[Detailed view of a decode \transition{}.\vspace*{-0.2cm}]{\includegraphics[width=\columnwidth]{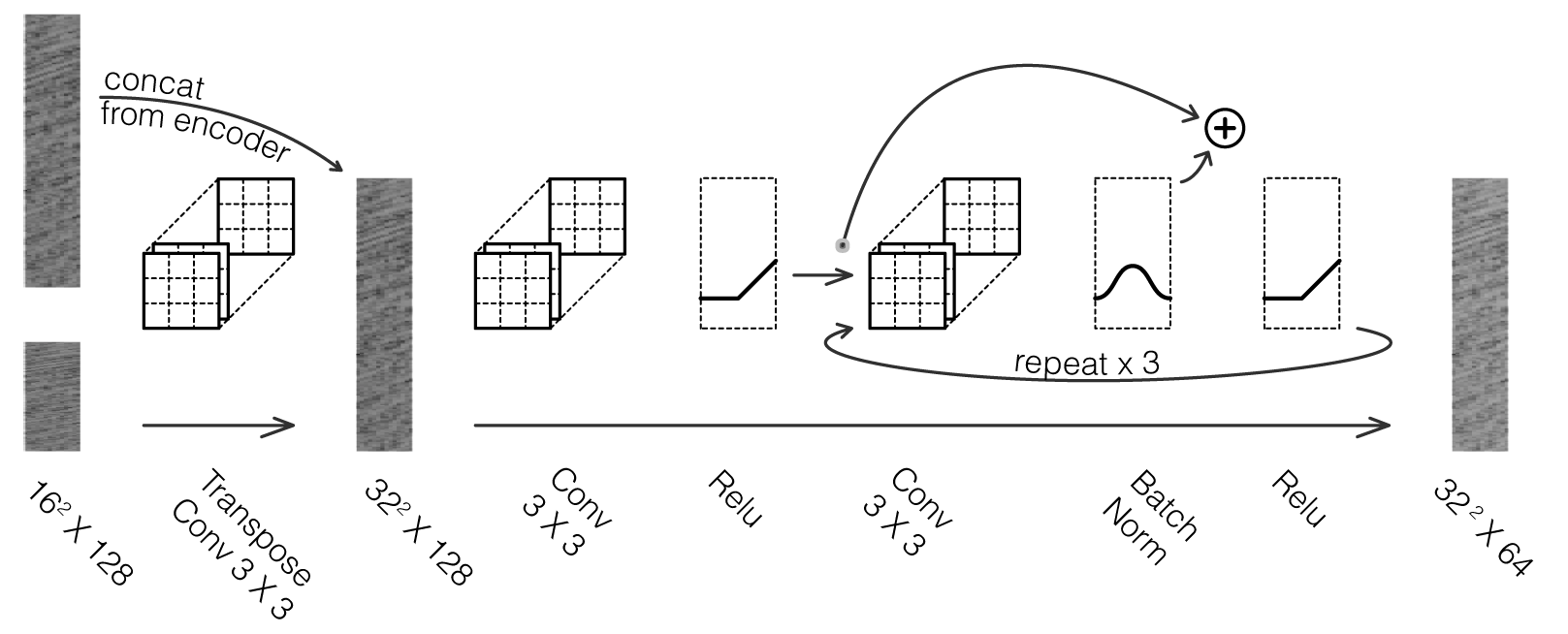}}
\caption{Detailed network architecture.}
\label{fig:unet_diagram}
\end{figure}

%% file: latex/results.tex
\section{Experiments}
\label{sec:exp}
\input{figures/wm_results/batch_example.tex}

\input{figures/wm_results/wm_results.tex}

\input{figures/wm_compare/wm_compare.tex}
\input{latex/Tables/watermark_compare.tex}

\input{latex/Tables/inpaint_table.tex}
\input{figures/inpaint/inpaint_compare.tex}
\input{figures/wm_results/real_water.tex}

We show the results of our method on various types of visual motifs: color and gray scale text,
emojis~\cite{emojidata}
and geometric shapes.
We have trained a different network for each type of visual motif. 
In all of our experiments, our background images consist of 2700 training and 300 separate testing  images, randomly chosen from the Microsoft COCO ‘val2014’ dataset \cite{lin2014microsoft}.
Our code is available at  \url{https://github.com/amirhertz/visual_motif_removal}.

\subsection{Transparent visual motifs}

{\bf Training.}
To train the network, we synthesize several sets of images embedded with a certain visual motif type (see Figure \ref{fig:batch_example}).
For each type, we have synthesized 20,000 $512 \times 512$ corrupted images as follows:  

\noindent (i) Colored texts with Latin characters that have random font and color. These are blended into the background images using a constant opacity $\alpha\in\left(0.3, 0.7\right)$ within each image. \\
(ii) Gray scale texts are characterized by random light color and dark borders. They are embedded with a uniformly distributed opacity blending field with 10\% variance around the constant $\alpha$ and a random spatial perturbations of up to one pixel shifts. \\
(iii) The emojis dataset has $800$ randomly chosen objects. They are blended with a constant opacity in the range $\alpha\in\left(0.4, 0.6\right)$ and are applied with the same perturbations and 10\% opacity variance. \\
(iv) Geometric grayscale shapes such as lines, rectangles and ellipses; they are embedded with a constant opacity in the range $\alpha\in\left(0.2, 0.9\right)$.

For each training image, we embed up to $10$ visual motifs using random positions, scales, crops and rotations for data augmentation.
To speed up training, we use $128 \times 128$ sized patches randomly selected from the corrupted image as input to the network (see Figure \ref{fig:batch_example}), but test on full resolution images.

{\bf Testing.}
The test images consist of full resolution images, which are unseen during training.
For the text experiments, we trained on Latin characters, but the visual motifs are unseen characters from other languages, \emph{e.g.,} Chinese, Japanese, Hindi, or Georgian.
The emojis test set consists of emojis that have not been used during training. Various reconstruction results are shown in Figures \ref{fig:teaser} and \ref{fig:wm_results}, as well as in the supplementary material.
To highlight the difference between the test and train sets, we show several test emojis (\emph{e.g.,} the ones that appear in figures \ref{fig:teaser} and \ref{fig:wm_results}) with their $5$ Euclidean nearest neighbors in the supplementary material.

{\bf Single visual motif removal.}
We compare the performance of our method  to four algorithms:  closed-form solution to natural image matting (CFM) \cite{levin2008closed}, multi-image reconstruction (MMR) \cite{dekel2017effectiveness} and two deep image reflection separation approaches (SIRF, BDN) \cite{Yang2018Seeing,Zhang2018SingleIR}, which we adapt to the motif removal problem. To facilitate the task of MMR, we use only a single unseen emoji as the overlaid visual motif on the entire test set (1000 images in total). To facilitate the task for CFM, we provide the ground-truth motif blending image.

Our test images are divided into three groups with increasing levels of matting deformation. In the first group, the visual motif has a fixed size of $160 \times 160$ px. It appears at a random position and blended with a constant opacity within each image $\alpha \in \left(0.4, 0.6\right)$. For the second group, we apply the blending with added perturbations to the edges of the visual motif and a varying opacity. For the third group, variations in size and rotation were added. The visual motif's size was $120 \times 120 - 200 \times 200$ px and randomly rotated between $\theta \in \left(-20^{\circ}, 20^{\circ} \right)$.

The results are measured by comparing the reconstructed images to the ground truth one under PSNR and the structural similarity index (SSIM)~\cite{wang2004image}.
As can be seen in both Table \ref{tab:watemark_compare} and the sample images from the second test in figure \ref{fig:wm_compare}, our method performs better both in the easier and more challenging scenarios. More visual results of all tests are displayed in the supplementary material.

Note that non-deep approaches do not handle the more difficult scenario (our method is not effected). In addition, the MMR algorithm has a missed detection rate of $4.5\%$ (the location of the visual motif is not correctly identified). These cases are excluded from the overall results.

The reflection separation works follow an \emph{encoder-decoder} framework, since they must generalize to reflections potentially present in the entire image. The major difference between reflection and visual motif removal in images is the distribution of transparent pixels. 
Motif embedded images will always contain content from the \emph{background} image. 
Since our method specializes in motifs, we constrain the reconstruction to the (learned) binary motif mask. 
In our ablation study (see section ~\ref{sec:abl}), we demonstrated that using a regular auto-encoder is inferior to a network with a motif mask branch.


{\bf Watermark removal.}
Naturally, our method can be \emph{abused}, and be used for removing watermarks from protected images. 
See Figure \ref{fig:real_water} for examples of removing watermarks from various stock photography services. More examples are shown in the supplementary material.

For this task, we trained the network on a combination of the different training sets outlined above. Specifically, we used the synthetic light texts set (ii) and the gray scale shapes set (iv), in addition to a dataset of synthetic white texts. These texts were embedded using a constant opacity $\alpha\in\left(0.2, 0.7\right)$, followed by a $3 \times 3$ Gaussian  blur filter, which causes a gradual matting between the texts and the background images.

\subsection{Inpainting}

Another type of visual motif we tested our method on is blind inpainting tasks: the network has no explicit priors on the background image or the mask of the corrupted regions.
We evaluate our network on two levels of inpainting regions. The first is composed from $256 \times 256$ pixels images on which we tiled random black texts of the font Helvetica light and with random range of rotations and sizes between $80$ to $120$ pixels per word.
For the second experiment, we used a bolder Helvetica medium font with up to $150$ pixels per word. 
For this task, we have trained our baseline configuration without the $Vm$ branch since the retrieval of the pixel-mask is equivalent to the retrieval of the black regions. 
Our network was trained on the same $2700$ images which were randomly cropped to $64 \times 64$ pixels patches.
Each test dataset contains 100 unseen images and unseen words.
We compared our method with state-of-the-art methods \cite{ren2015shepard, roth2009fields, zoran2011learning}, which unlike our method, were supplied with the ground truth masks of the opaque regions.
Table \ref{tab:inpaint_table} summarizes the quantitative results on the 100 test images of each experiment.
Figure \ref{fig:inpaint_compare} shows visual comparisons to the other methods. The results clearly demonstrate that our method performs better in the task of completing real imagery patterns and achieves sharper results when filling large corrupted regions.

We also used our pre-trained network on the sample images from these methods (see supplementary material). Our method still outperforms even when using their sample images.

\subsection{Ablation study}
\label{sec:abl}
\input{latex/Tables/ablation_nets_table.tex}
\input{latex/Tables/ablation_params.tex}

In our ablation study, we verify our network configurations by testing the quality of the results obtained by the different branches configurations and the inner architecture setup.

We generate the training dataset for the study as described previously, using emojis~\cite{emojidata} as the visual motifs blended on images of size $256 \times 256$ pixels. The motif size is uniformly randomly sampled in the range $(80, 160)$ pixels. We blend the motif with a blending constant $\alpha$ which is randomly sampled $\in (0.3, 0.7)$.
Each tested network is trained for 100 epochs on patches of size $128 \times 128$.
We test the performance of the networks on 200 unseen images blended with unseen motifs.

The network configurations we compared include an auto-encoder U-Net model, our baseline two branches model, a three branches model with a separated branch for the motif (Baseline + Vm) and the chosen model (Baseline + shared Vm) with three branches as well as shared weight between the image and the motif decoders. 

Table \ref{tab:ablation_nets_table} summarizes the quantitative results on the test set (visual comparisons in the supplementary material). Both results show that the auto-encoder network, without the mask branch, cannot separate the motif from the background image. The chosen model with the shared motif decoder (Baseline + shared Vm) generates the best restoration within the corrupted regions.

In addition, we explore the impact of the size of the shared weights between the image and the motif branches and try shallower and deeper setups by changing the number of residual blocks within each encoder-decoder \transition{}. 
The quantitative results for these studies in Table \ref{tab:ablation_table_params} show that the addition of (up to three) res-blocks and the use of the shared weights (up to two decode levels) improve the reconstruction of the image.




%% file: figures/wm_results/batch_example.tex
\begin{figure}
\includegraphics[width=\columnwidth]{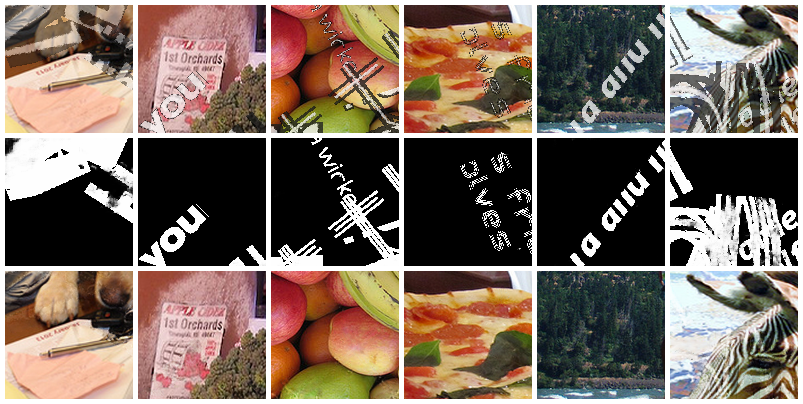}
\vspace*{-0.6cm}
\caption{{\em Training patches}. First row contains $128 \times 128$ synthesized corrupted patches. The middle row shows the output of the mask decoder (estimated masks) and the bottom row presents the final reconstructed patches.}
\label{fig:batch_example}
\end{figure}

%% file: figures/wm_results/wm_results.tex
\begin{figure*}
\includegraphics[width=\textwidth]{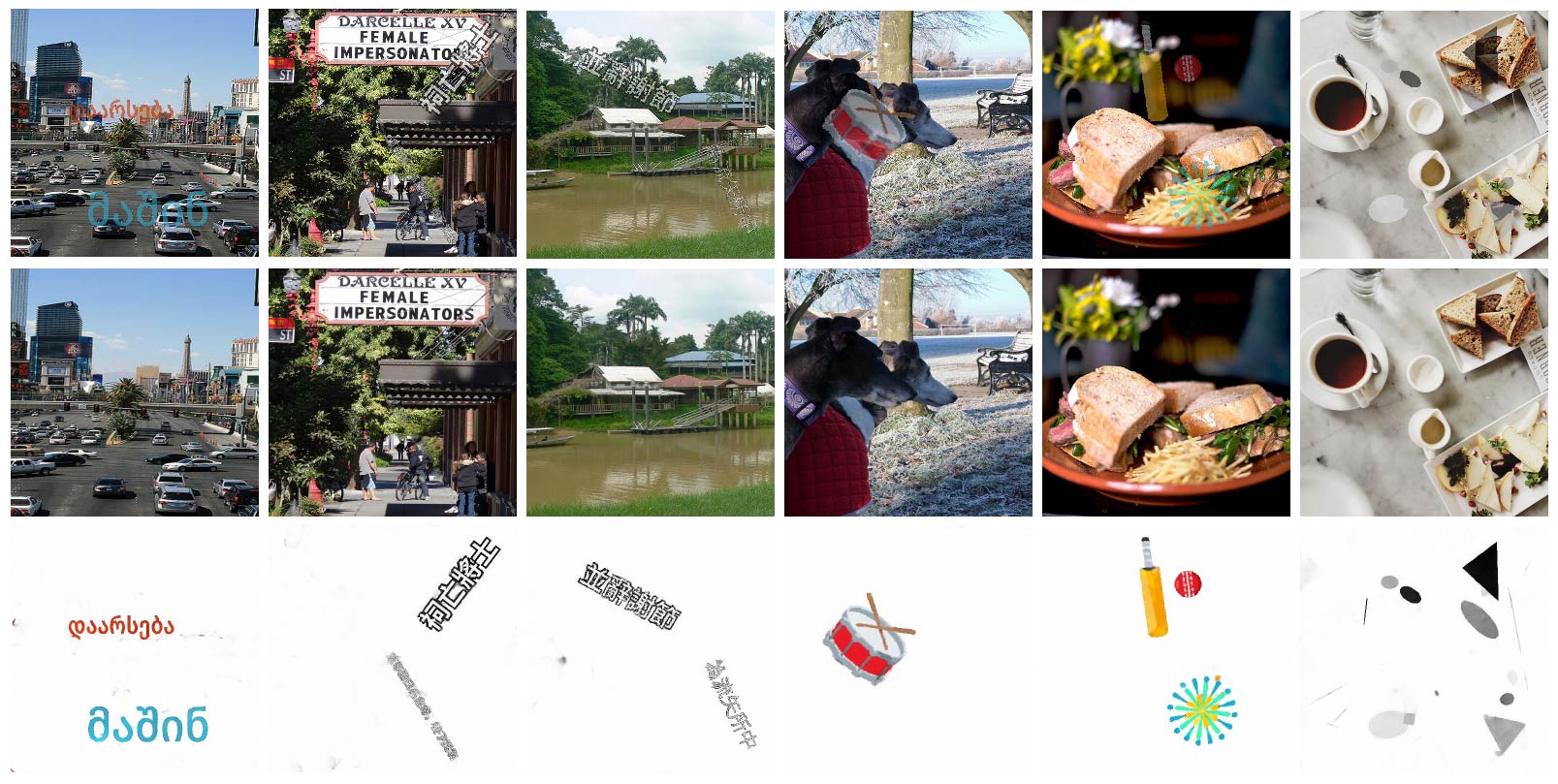}
\vspace*{-0.6cm}
\caption{Semi-transparent visual motif removal. Top: test images embedded with unseen motifs. Middle: our reconstructed background images. Bottom: our reconstructed visual motifs. More results are presented in the supplementary material.}
\label{fig:wm_results}
\end{figure*}

%% file: figures/wm_compare/wm_compare.tex
\newcolumntype{C}[1]{>{\let\newline\\\arraybackslash\hspace{0pt}}m{#1}}

\begin{figure*}
\newcommand{\tfig}{2.5}
\begin{center}
\renewcommand{\arraystretch}{0.7}
\begin{tabular}{c}

\includegraphics[width=\textwidth, trim={0 2.1cm 0 0},clip]{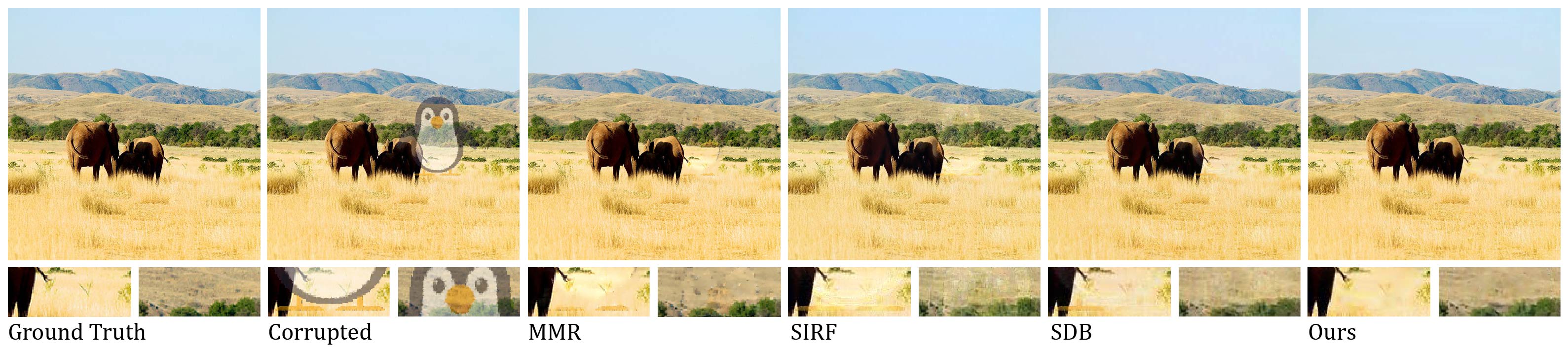} \\

\begin{tabular}{C{\tfig cm} C{\tfig cm} C{\tfig cm} C{\tfig cm} C{\tfig cm} C{\tfig cm}}

{\footnotesize Ground Truth} & {\footnotesize Corrupted} & {\footnotesize MMR} & {\footnotesize SIRF} & {\footnotesize SDB} & {\footnotesize Ours}\\
\end{tabular}
\end{tabular}
\end{center}
\vspace*{-12pt}
\caption{Visual comparisons for the second test in Table~\ref{tab:watemark_compare} for removing a single watermark from a multi-image collection. More results may be found at the supplementary material.}
\label{fig:wm_compare}
\end{figure*}

%% file: latex/Tables/watermark_compare.tex
\newcolumntype{?}{!{\vrule width 1pt}}
\begin{table}
\noindent\begin{tabular*}{\columnwidth}{l ?  c ? c  ? c }
    \toprule
    &
    Translation & Pert. + Opc. & Scale + Rot. \\
    & \small{PSNR / SSIM} & \small{PSNR / SSIM} & \small{PSNR / SSIM} 
    \\
    \hline
    \hspace{-0.07in} \small{CFM}\cite{levin2008closed} \hspace{-0.07in} & 24.16 / 0.976 &  N/A &  N/A \\
    \hline
    \hspace{-0.07in} \small{MMR}\cite{dekel2017effectiveness} \hspace{-0.07in} & 37.41 / 0.977 & 33.07 / 0.966 &  N/A \\
    \hline
    \hspace{-0.07in} \small{SDB}\cite{Yang2018Seeing} \hspace{-0.07in} & 34.64 / 0.973 & 34.82 / 0.972 &  34.69 / 0.972 \\
    \hline
     \hspace{-0.07in} \small{SIRF}\cite{Zhang2018SingleIR} \hspace{-0.07in} & 31.18 / 0.970 &  32.87 / 0.970 &  32.90 / 0.969 \\
    \hline
    \textbf{Ours} & \textbf{38.46} / \textbf{0.986} & \textbf{38.08} / \textbf{0.986} & \textbf{37.63} / \textbf{0.983}\\
    \bottomrule
\end{tabular*}
\vspace*{-0.2cm}
\caption{Watermark removal comparison to existing approaches when applied on watermarks with perturbations to translation, scale and rotation.}
\label{tab:watemark_compare}

\end{table}

%% file: latex/Tables/inpaint_table.tex
\newcolumntype{?}{!{\vrule width 1pt}}
\begin{table}
\noindent\begin{tabular*}{\columnwidth}{l ?  c ? c }
\toprule
& \begin{tabular}[c]{@{}l@{}}Light Font Test\\ PSNR / SSIM\end{tabular} & \begin{tabular}[c]{@{}l@{}}Bold Font Test\\ PSNR / SSIM\end{tabular} \\ \hline
Sh-CNN\cite{ren2015shepard}  & 31.879 / 0.9522                                                      & 28.436 / 0.9118                                                     \\ \hline
FoE\cite{roth2009fields}    & 38.155 / 0.9886                                                      & 33.360 / 0.9675                                                     \\ \hline
EPLL\cite{zoran2011learning}   & 38.672 / 0.9884                                                      & 33.377 / 0.9675                                                     \\ \hline
\textbf{Ours}   & \textbf{39.079} / \textbf{0.9890}                                                    & \textbf{34.676} / \textbf{0.9710}                                        \\ \bottomrule
\end{tabular*}
\vspace*{-0.1cm}
\caption{Quantitative inpainting results for images corrupted by lighter and bolder font.}
\label{tab:inpaint_table}
\vspace*{-0.1cm}
\end{table}

%% file: figures/inpaint/inpaint_compare.tex

\newcolumntype{C}[1]{>{\let\newline\\\arraybackslash\hspace{0pt}}m{#1}}

\begin{figure*}
\newcommand{\tfig}{2.5}
\begin{center}
\renewcommand{\arraystretch}{0.7}
\begin{tabular}{c}

\includegraphics[width=\textwidth, trim={0 1cm 0 0},clip]{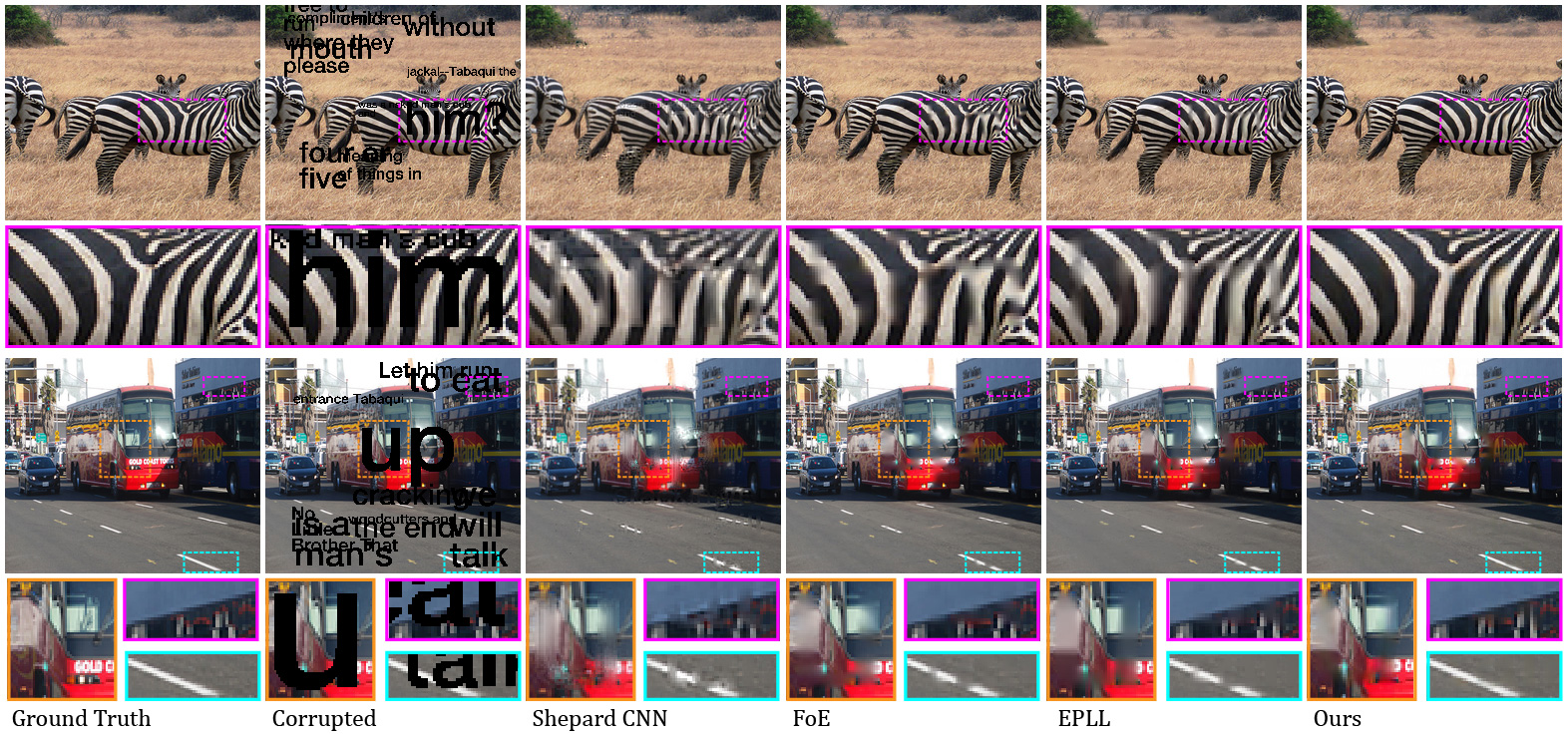} \\

\begin{tabular}{C{\tfig cm} C{\tfig cm} C{\tfig cm} C{\tfig cm} C{\tfig cm} C{\tfig cm}}

{\footnotesize Ground Truth} & {\footnotesize Corrupted} & {\footnotesize Shepard CNN} & {\footnotesize FoE} & {\footnotesize EPLL} & {\footnotesize Ours}\\

\end{tabular}
\end{tabular}
\end{center}
\vspace*{-12pt}
\caption{Visual comparisons with other inpainting methods. Our results are less blurred and the patterns distracted by the letters are inpainted in a more continuous way.}
\label{fig:inpaint_compare}
\end{figure*}

%% file: figures/wm_results/real_water.tex
\begin{figure*}
\includegraphics[width=\textwidth]{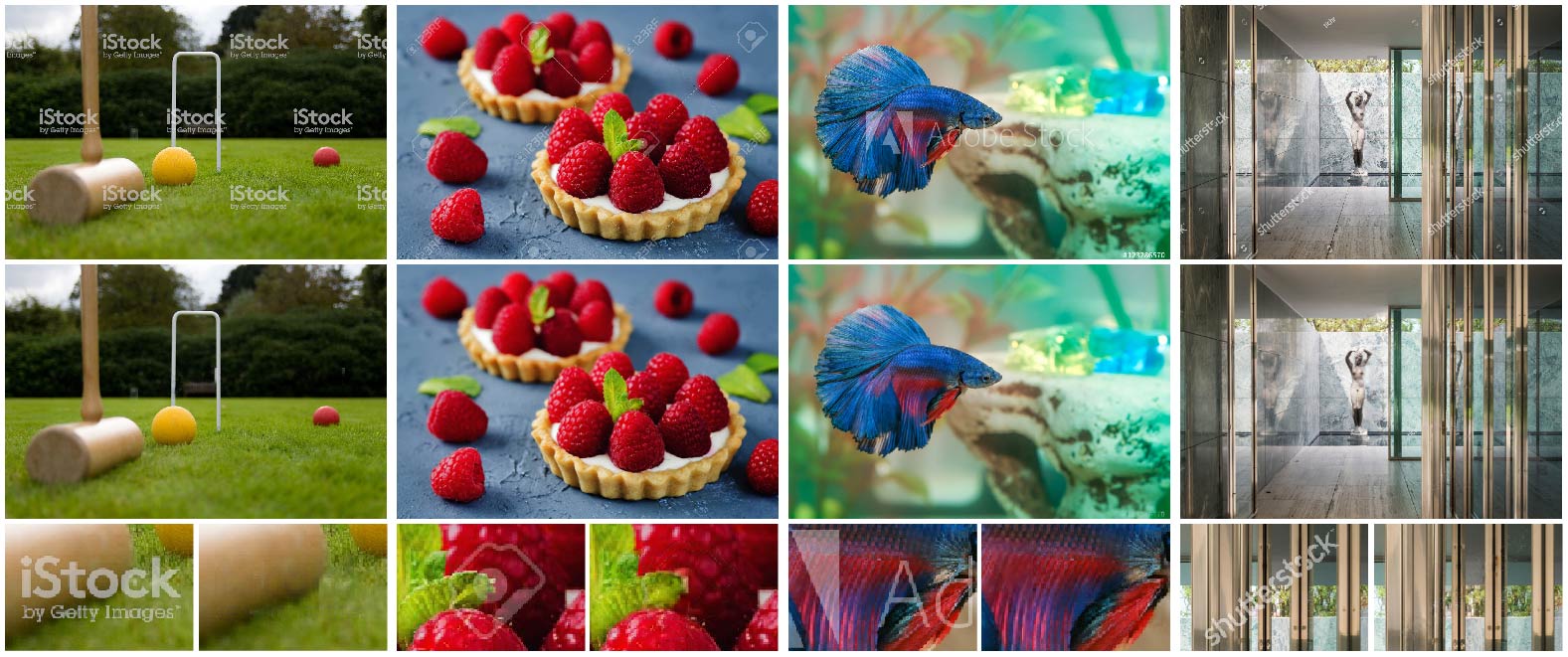}
\vspace*{-0.5cm}
\caption{{\em Watermark removal examples}. Top: input images, middle: reconstruction results, and bottom: enlarged patches.}
\label{fig:real_water}
\vspace*{-0.3cm}
\end{figure*}

%% file: latex/Tables/ablation_nets_table.tex
\begin{table}
\noindent\begin{tabular*}{\columnwidth}{c | c | c | c | c}
    \toprule
     &
     \begin{tabular}{@{}c@{}}Auto- \\ encoder\end{tabular} &
     Baseline &
     \begin{tabular}{@{}c@{}}Baseline \\ + VM\end{tabular} &
     \begin{tabular}{@{}c@{}}\textbf{Baseline} + \\ \textbf{shared VM}\end{tabular} \\
    \hline
     PSNR & 27.729 & 37.359  & 37.564  & \textbf{37.636} \\
     \hline
     SSIM & 0.9443 & 0.9882 & 0.9883 & \textbf{0.9884}\\
    \bottomrule
\end{tabular*}
\vspace*{-0.2cm}
\caption{Ablation study of the network components.}
\label{tab:ablation_nets_table}
\end{table}

%% file: latex/Tables/ablation_params.tex
\newcolumntype{?}{!{\vrule width 1pt}}
\begin{table}
\noindent\begin{tabular*}{\columnwidth}{r | c ? r| c }
    \toprule
    \begin{tabular}{@{}r@{}} \# res \\ blocks\end{tabular} &
    PSNR / SSIM &
    \begin{tabular}{@{}r@{}} Shared \\ depth \end{tabular} &
    PSNR / SSIM \\
    \hline
     1 block & 35.964 / 0.9848 & 1 level  & 37.414 / 0.9883   \\
     \textbf{3 blocks} & \textbf{37.636 / 0.9884} & \textbf{2 levels}  & \textbf{37.636 / 0.9884}   \\
     5 blocks & 37.411 / 0.9880 & 3 levels  & 37.4387 / 0.9882  \\
    \bottomrule
\end{tabular*}
\vspace*{-0.2cm}
\caption{Ablation study of the network depth.}
\label{tab:ablation_table_params}
\vspace*{-0.2cm}
\end{table}

%% file: latex/conclusion.tex
\section{Conclusion}

We have presented a method for the identification and removal of visual motifs embedded in images. Our method is the first to remove visual motifs from images blindly (\emph{i.e.,} without explicit prior information). It should be stressed that unlike previous methods, our method does not require any user input or making intricate assumptions about the visual motif.

Deep neural networks are capable of generalizing beyond the examples presented during training. This enables us to train on one set of visual motifs, and during test time remove a large variety of other visual motifs, which may differ significantly in size, location, colors and even shape from the training set. 
This generalization power is clearly demonstrated in the successful removal of Indian, Chinese and Japanese characters when training on Latin characters (\emph{e.g.,} Figure~\ref{fig:teaser} and~\ref{fig:wm_results}).

A notable advantage of our approach is the addition of a decoder network for reconstructing the distracting visual motif. 
We have shown that this leads to a better latent image reconstruction and visual motif detection. 
This is particularly notable in the removal of semi-transparent visual motifs, which are embedded with varying transparency or small spatial perturbations.

\input{figures/wm_results/failure.tex}

When the visual motif is embedded over a uniform background, the motif traces are still somewhat noticeable (see example in Figure~\ref{fig:failure}).
We believe this limitation can be alleviated by training our network using hard example mining to make it more accurate in cases where the background of the visual motif is relatively uniform. Another interesting future direction would be to add a generative network for synthesizing complex visual motifs, such as raindrops or other visual distractors.




%% file: figures/wm_results/failure.tex
\begin{figure}
\includegraphics[width=\columnwidth]{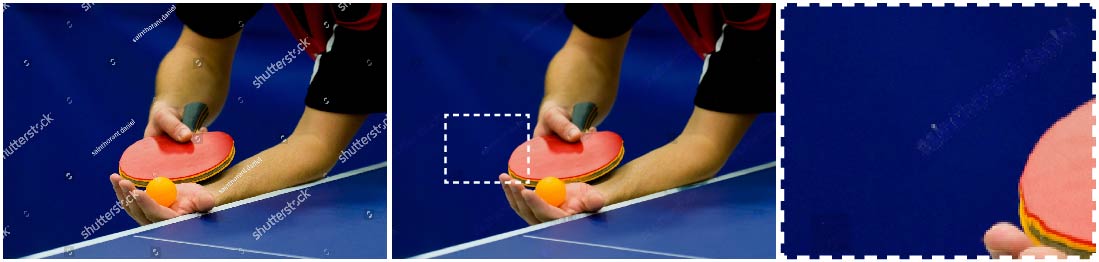}
\vspace*{-0.5cm}
\caption{{\em Failure case}. Our method may struggle when the motif is embedded in flat uniform regions.}
\label{fig:failure}
\vspace*{-0.3cm}
\end{figure}